\documentclass[%
 reprint,
 showpacs,
 aps,
 superscriptaddress,
 pra,]{revtex4-2}

\usepackage{hyperref} %Figure用に追加
\usepackage{bm}% bold math
\usepackage{amsmath}
\usepackage{amssymb}
\usepackage{amsfonts}
\usepackage{braket}
\usepackage[dvipdfmx]{graphicx}
\usepackage{color}
\usepackage{here}
\usepackage{enumitem}
\usepackage{array}
\usepackage {diagbox}
\usepackage{subcaption}

\begin{document}
\title{Efficient Bit Labeling in Factorization Machines with Annealing for Traveling Salesman Problem}

\author{ Shota Koshikawa, Aruto Hosaka, and Tsuyoshi Yoshida \\ \thanks{\small \itshape Information Technology R\&D Center, Mitsubishi Electric Corporation, Kanagawa 247-8501, Japan}}
%\affiliation{%
%Mitsubishi Electric Corporation, Information Technology R\&D Center, Kanagawa 247-8501, Japan
%}%

\date{\today}

\begin{abstract}
  To efficiently find an optimum parameter combination in a large-scale problem, it is a key to convert the parameters into available variables in actual machines. Specifically, quadratic unconstrained binary optimization  problems are solved with the help of machine learning, e.g., factorization machines with annealing, which convert a raw parameter to binary variables. This work investigates the dependence of the convergence speed and the accuracy on binary labeling method, which can influence the cost function shape and thus the probability of being captured at a local minimum solution. By exemplifying traveling salesman problem, we propose and evaluate Gray labeling, which correlates the Hamming distance in binary labels with the traveling distance. Through numerical simulation of traveling salesman problem up to 15 cities at a limited number of iterations, the Gray labeling shows less local minima percentages and shorter traveling distances compared with natural labeling.
\end{abstract}

\maketitle

\section{Introduction}

Combinatorial optimization problems have gained significant attention across various domains, including logistics, transportation systems, and manufacturing \cite{transportation, manufacturing}, due to their wide-range applications and potentials for cost reduction and efficiency improvement. The computational complexity of these problems is generally classified to NP hardness, resulting in substantially challenging to approach the optimal solution at a reasonable amount of computational resource \cite{NP_survey, NP_graph}. Renowned for its computational complexity as an NP-hard problem, the traveling salesman problem (TSP) serves as a cornerstone in numerous fields, and being vigorously researched \cite{neuro_ising, SB_TSP}. 

The complexity of such difficult problems can be relaxed by combining machine learning. Especially, factorization machines with annealing (FMA) \cite{FMA_META, FMA_ENC, FMA_PHOTONIC, bbo_kadowaki, bbo_matsumori, FMA_MTJ} is a useful technique for black-box optimization \cite{BBO, BBO_FMQA_REVIEW, bbo_doi, bbo_cross_entropy, bbo_izawa}. FMA employs factorization machines (FM) \cite{FM} with binary variables as a surrogate model. Since the model takes the form of a quadratic unconstrained binary optimization (QUBO), Ising machines can be utilized to efficiently find a good solution for the model \cite{lucas}. 

The performance of a QUBO solver depends on the labeling method, i.e., how the actual nonbinary variables are replaced by binary variables available in the solver. While the labeling method is a key to characterize how frequently the solver is captured at local solutions, there has been limited research on it \cite{QUBIT_ENCODING, TSP_VQE, kikuchi}. 
It aims at creating a smoother energy landscape by assigning bit states with short Hamming distances to binary variable configurations close in the solution space. By ensuring that similar solutions are represented by bit states with short Hamming distances, we hypothesize that we can achieve more efficient optimization.

According to the situation described above, this work originally contributes on QUBO formulation of TSP with reduced number of bits by employing FMA, proposal of Gray labeling useful for avoiding local solutions based on the idea of \emph{similar bits for similar routes}, proposal of the metric for local solution characterization, and comparison of conventional natural labeling and Gray labeling.

The remainder of this paper is structured as follows: based on the preliminaries of FMA and TSP in Sec.~\ref{sec:pre}, Sec.~\ref{sec:BLM} explains bit labeling methods of natural and Gray labeling. Sec.~\ref{sec:LS} then introduces a local solution metric for efficient characterization of QUBO problems. To validate our approach, Sec.~\ref{sec:NS} performs numerical simulations of FMA-based TSP solvers with two labeling methods. Finally, Sec.~\ref{sec:conc} concludes the paper.

\section{Preliminaries} \label{sec:pre}
This section reviews fundamentals of FMA and TSP.

\subsection{Factorization machines with annealing} \label{sec:FMA}
Rendle proposed an FM model for high prediction performance with efficient high-order feature interactions. The prediction is given by the sum of the linear and the quadratic-order interaction terms \cite{FM}: 

\begin{equation}
    \begin{aligned}
 %     y = w_0 +\sum_{i=1}^{\mathsf{n}}{w_ix_i} + \sum_{i=1}^{\mathsf{n}}\sum_{j=i+1}^{\mathsf{n}} \langle\mathbf{v}_i, \mathbf{v}_j\rangle x_i x_j .
      y = w_0 +\sum_{\mathsf{i}=1}^{\mathsf{n}}{w_\mathsf{i}x_\mathsf{i}} + \sum_{\mathsf{i}=1}^{\mathsf{n}}\sum_{\mathsf{j}=\mathsf{i}+1}^{\mathsf{n}} \langle\mathbf{v}_\mathsf{i}, \mathbf{v}_\mathsf{j}\rangle x_\mathsf{i} x_\mathsf{j} .
    \end{aligned}
  \end{equation}
The input data is represented as a feature vector $\mathbf{x} = (x_1, x_2, \ldots , x_\mathsf{n})$ of $\mathsf{n}$ real-valued features, and $y$ is an objective variable. $w_0$ is the global bias, $w_\mathsf{i}$ is the weight of the $\mathsf{i}$-th feature, and a weight vector $\mathbf{w} = (w_1, \cdots, w_\mathsf{n})$. $\mathbf{v}_\mathsf{i}$ is the $\mathsf{k}$-dimensional latent vector of the $\mathsf{i}$-th feature, and the vector sequence $\mathbf{V} = (\mathbf{v}_1, \cdots, \mathbf{v}_\mathsf{n})$. The interaction between features $x_\mathsf{i}$ and $x_\mathsf{j}$ is approximated by the inner product $\langle\mathbf{v}_\mathsf{i}, \mathbf{v}_\mathsf{j}\rangle$. 
The model parameters $(w_0, \mathbf{w}, \mathbf{V})$ are optimized to minimize the error between the predicted and actual values on the training data.

Unlike support vector machines, FM use factorized parameters to model all variable interactions. In traditional polynomial models, it was necessary to prepare individual interaction parameters for each combination, such as $w_{\mathsf{i}\mathsf{j}}x_\mathsf{i}x_\mathsf{j}$. However, $x_\mathsf{i} x_\mathsf{j}$ becomes mostly zero in sparse data, making it almost impossible to calculate $w_{\mathsf{i}\mathsf{j}}$. In contrast, FM represent the magnitude of the interaction of $x_\mathsf{i}x_\mathsf{j}$ as $\langle\mathbf{v}_\mathsf{i}, \mathbf{v}_\mathsf{j}\rangle$, that is, no longer mutually independent of each $w_{\mathsf{i}\mathsf{j}}$. Therefore, even if one or both of the interaction components are zero, if there is a non-zero component of $x_\mathsf{i}$ or $x_\mathsf{j}$ somewhere, the parameters $\mathbf{v}_\mathsf{i}$ and $\mathbf{v}_\mathsf{j}$ can be learned. This implies that FM can indirectly learn interaction effects even from data without the target interaction components. Thus FM are robust in handling sparse data and have a relatively low computational cost \cite{FM}. This makes it useful for high-dimensional sparse data applications.

FM can be combined with an optimization method of annealing \cite{FMA_META, FMA_ENC, FMA_PHOTONIC, bbo_kadowaki, bbo_matsumori, FMA_MTJ}, where the combination is called FMA. The model equation of FM with binary variables can be rewritten in the QUBO form:

\begin{equation}
  y = w_0 + \sum_{\mathsf{i}=1}^{\mathsf{n}}\sum_{\mathsf{j}=\mathsf{i}}^{\mathsf{n}} Q_{\mathsf{i}\mathsf{j}}x_\mathsf{i}x_\mathsf{j} ,
\end{equation}
where $Q = (Q_{\mathsf{i}\mathsf{j}})$ is an $\mathsf{n} \times \mathsf{n}$ QUBO matrix, $Q_{\mathsf{i}\mathsf{i}}= w_\mathsf{i}$, $Q_{\mathsf{i}\mathsf{j}}= \langle\mathbf{v}_\mathsf{i}, \mathbf{v}_\mathsf{j}\rangle$. Now we explain the optimization method for black-box optimization problems using FMA. The FMA approach comprises four main phases that are repeated in an iterative cycle \cite{FMA_META}:

\begin{itemize}

   \item \textbf{Training}: The FM model is trained using the available training data. A solution candidate of the single bit sequence $\mathbf{b}$ were randomly generated, and the pairs of $\mathbf{b}$ and corresponding energy (objective variables) were added for the initial training. The parameters of the FM are optimized to minimize the mean-squared error between the predicted values and the actual energy values.    
   \item \textbf{Sampling}: New bit sequences are generated from the trained FM model, focusing on samples with low predicted energy values. Since the FM model is formulated as a QUBO, quantum or classical annealing techniques can be employed to find low-energy states, which correspond to good samples.
   
   \item \textbf{Conversion}: The bit sequences generated in the sampling are converted back to the original optimization problem's parameters. This aspect will be detailed in Sec.~\ref{sec:BLM}.

   \item \textbf{Evaluation}: The costs are simulated or experimented using parameters obtained at the previous iteration, and the pairs of the binarized parameters and the corresponding energy are used to update the training data.
\end{itemize}

The FMA approach iterates through these four phases multiple times, gradually refining the approximation of the black-box function, in this case QUBO, and improving the quality of the solutions. After a given number of iterations, the best sample found during the optimization process is returned as the final solution.

\subsection{Traveling salesman problem} \label{sec:TSP}

\begin{table*}[t]
  \begin{center}
    \caption{Coordinates of TSP cities $( \alpha, \beta )$ in this work.}
    \begin{tabular}{|c|c|c|c|c|c|c|} \hline 
      $i$ & $N=5$ & $N=7$ & $N=9$ & $N=11$ & $N=13$ & $N=15$ \\ \hline
       0 & (0.069, 0.530) & (0.865, 0.693) & (0.961, 0.983) & (0.235, 0.339) & (0.561, 0.048) & (0.795, 0.361) \\ \hline
 1 & (0.204, 0.891) & (0.266, 0.285) & (0.598, 0.990) & (0.895, 0.135) & (0.828, 0.879) & (0.743, 0.529) \\ \hline
 2 & (0.531, 0.034) & (0.436, 0.059) & (0.080, 0.916) & (0.241, 0.817) & (0.081, 0.271) & (0.352, 0.303) \\ \hline
 3 & (0.837, 0.204) & (0.051, 0.984) & (0.511, 0.200) & (0.995, 0.728) & (0.244, 0.897) & (0.192, 0.074) \\ \hline
 4 & (0.695, 0.688) & (0.861, 0.032) & (0.468, 0.734) & (0.432, 0.641) & (0.863, 0.387) & (0.472, 0.679) \\ \hline
 5 & - & (0.271, 0.592) & (0.980, 0.059) & (0.605, 0.838) & (0.543, 0.096) & (0.399, 0.021) \\ \hline
 6 & - & (0.990, 0.267) & (0.643, 0.676) & (0.999, 0.371) & (0.032, 0.222) & (0.777, 0.101) \\ \hline
 7 & - & - & (0.096, 0.167) & (0.283, 0.926) & (0.686, 0.991) & (0.990, 0.425) \\ \hline
 8 & - & - & (0.026, 0.378) & (0.504, 0.065) & (0.661, 0.484) & (0.869, 0.470) \\ \hline
 9 & - & - & - & (0.982, 0.150) & (0.246, 0.295) & (0.782, 0.662) \\ \hline
10 & - & - & - & (0.673, 0.783) & (0.047, 0.608) & (0.614, 0.460) \\ \hline
11 & - & - & - & - & (0.381, 0.031) & (0.109, 0.430) \\ \hline
12 & - & - & - & - & (0.773, 0.593) & (0.000, 0.035) \\ \hline
13 & - & - & - & - & - & (0.427, 0.148) \\ \hline
14 & - & - & - & - & - & (0.395, 0.843) \\ \hline
    \end{tabular}
    \label{tab:coordinates}    
  \end{center}
\end{table*}

  TSP is one of the most widely studied combinatorial optimization problems \cite{applegate, laporte}, which tries to find the shortest route that visits all predefined points exactly once and returns to the origin. This can be extended to various optimization problems, such as the component assembly sequence in manufacturing, delivery routes in logistics, and data transmission paths in telecommunication networks.

  Regarding the complexity of TSP, as increasing the number of cities $N$, the total number of possible routes grows exponentially and reaches $(N-1)!$, e.g., $8.7 \times 10^{10}$ routes for $N=15$. It is impractical to perform brute-force search under a case with large $N$.
  Various algorithms have been proposed to find the optimal solution for TSP, including well-known dynamic programming and branch-and-bound algorithms, reaching the exact solution \cite{held_karp, bellman}. One of those, Held-Karp algorithm \cite{held_karp} shows the time complexity of $O(N^2 2^N)$. On the other hand, these algorithms are difficult to apply to a case with large $N$, thus often combined with an approximation method, e.g., greedy algorithm \cite{TSP_greedy}, local search method \cite{TSP_local}, genetic algorithm \cite{TSP_GA}, ant colony optimization \cite{TSP_ant}, and quantum/simulated annealing \cite{TSP_QA}.
  
  In this work, $N=5$--$15$ cities are placed in rectangular coordinates $(\alpha , \beta)$, where $\alpha$ and $\beta$ ($\in [0,1]$) are randomly obtained as shown in Tab.~\ref{tab:coordinates}. Each city has a unique integer index $i \in \{ 0, 1, \ldots, N-1 \}$. The departure and destination city is indexed by 0. An arbitrary route is described as $\mathbf{r} = (r_1, r_2, \cdots, r_{N-1})$ except for the 0-th city. The objective is to minimize the distance:

  \begin{equation}
    \begin{aligned}
      d(\mathbf{r}) = \sum_{j=0}^{N-1}{\sqrt{(\alpha_{r_{j+1}} - \alpha_{r_{j}})^2 + (\beta_{r_{j+1}} - \beta_{r_{j}})^2}},
      \label{eq:dist}
    \end{aligned}
  \end{equation}
where $r_0=r_N=0$ according to the definition.

  \section{Bit labeling methods} \label{sec:BLM}
 This work treats TSP with FMA, so any variables in TSP must be redescribed by binary variables only. This section explains labeling methods of converting the TSP route $\mathbf{r}$ into the single bit sequence $\mathbf{b}$. In a well-known labeling method, $N^2$ bits are employed to formulate $N$-city TSP, resulting in a quadratic Hamiltonian \cite{lucas}. Recent works with improved labeling have reduced the number of bits to $N \mathrm{log}{N}$ \cite{ramezani, schnaus}. In this manuscript, $\mathrm{log}$ denotes the logarithm in base 2.

\begin{table*}[t]
  \begin{center}
    \caption{Examples of natural and Gray labeling. The forward labeling performs $\mathbf{r}\rightarrow\mathbf{b}$, and the inverse one does $\mathbf{\underline{b}}\rightarrow\mathbf{r}$. Each set of $\mathbf{\underline{m}}$, $\mathbf{\underline{b}}$, or $|\mathcal{\underline{S}}|$ is a superset of $m$, $\mathbf{b}$, and $|\mathcal{S}|$, respectively. Only the extended 8 elements are shown in the 3-rd, 5-th, 7-th, and 9-th columns.}
    \begin{tabular}{|c||c|c|c|c||c|c|c|c|} \hline
      route & \multicolumn{4}{c||}{Natural} & \multicolumn{4}{c|}{Gray} \\ \hline
      
      $\mathbf{r}$ & $m(=\underline{m})$ & $\underline{m}(\neq m)$ & $\mathbf{b} (= \underline{\mathbf{b}})$ & $\underline{\mathbf{b}} (\neq \mathbf{b})$ & $|\mathcal{S}| (= |\underline{\mathcal{S}}|)$ & $|\underline{\mathcal{S}}| (\neq |\mathcal{S}|)$ & $\mathbf{b} (= \underline{\mathbf{b}})$ & $\underline{\mathbf{b}} (\neq \mathbf{b})$ \\ \hline

(1,2,3,4) &	0 &	24 &	00000 &	11000 &	(0,0,0)  & (0,3,0) &	00000 &	01000 	\\ \hline
(1,2,4,3) &	1 &	25 &	00001 &	11001 &	(0,0,1)	 & (0,3,1) &	00001 &	01001 	\\ \hline
(1,3,2,4) &	2 &	26 &	00010 &	11010 &	(0,1,0)  & -	   &	00100 &	-     	\\ \hline
(1,3,4,2) &	3 &	27 &	00011 &	11011 &	(0,1,1)	 & -	   & 	00101 &	-     	\\ \hline
(1,4,2,3) &	4 &	28 &	00100 &	11100 &	(0,0,2)	 & (0,3,2) &	00011 &	01011	\\ \hline
(1,4,3,2) &	5 &	29 &	00101 &	11101 &	(0,1,2)  & -	   & 	00111 &	- 	\\ \hline
(2,1,3,4) &	6 &	30 &	00110 &	11110 &	(1,0,0)	 & (1,3,0) &	10000 &	11000	\\ \hline
(2,1,4,3) &	7 &	31 &	00111 &	11111 &	(1,0,1)	 & (1,3,1) &	10001 &	11001	\\ \hline
(2,3,1,4) &	8 &	-  &	01000 &  - &	(1,1,0)	 & -	   & 	10100 &	- 	\\ \hline
(2,3,4,1) &	9 &	-  &	01001 &	 - &	(1,1,1)	 & -	   & 	10101 &	- 	\\ \hline
(2,4,1,3) &	10 &	-  &	01010 &	 - &	(1,0,2)	 & (1,3,2) &	10011 &	11011	\\ \hline
(2,4,3,1) &	11 &	-  &	01011 &	 - &	(1,1,2)	 & -	   & 	10111 &	- 	\\ \hline
(3,1,2,4) &	12 &	-  &	01100 &	 - &	(0,2,0)	 & -	   & 	01100 &	-  	\\ \hline
(3,1,4,2) &	13 &	-  &	01101 &	 - &	(0,2,1)	 & -	   & 	01101 &	- 	\\ \hline
(3,2,1,4) &	14 &	-  &	01110 &	 - &	(1,2,0)	 & -	   & 	11100 &	- 	\\ \hline
(3,2,4,1) &	15 &	-  &	01111 &	 - &	(1,2,1)	 & -	   & 	11101 &	- 	\\ \hline
(3,4,1,2) &	16 &	-  &	10000 &	 - &	(0,2,2)	 & -	   & 	01111 &	- 	\\ \hline
(3,4,2,1) &	17 &	-  &	10001 &	 - &	(1,2,2)	 & -	   & 	11111 &	- 	\\ \hline
(4,1,2,3) &	18 &	-  &	10010 &	 - &	(0,0,3)	 & (0,3,3) &	00010 & 01010	\\ \hline
(4,1,3,2) &	19 &	-  &	10011 &	 - &	(0,1,3)	 & -	   & 	00110 &	- 	\\ \hline
(4,2,1,3) &	20 &	-  &	10100 &	 - &	(1,0,3)	 & (1,3,3) & 	10010 &	11010	\\ \hline
(4,2,3,1) &	21 &	-  &	10101 &	 - &	(1,1,3)	 & -	   & 	10110 &	- 	\\ \hline
(4,3,1,2) &	22 &	-  &	10110 &	 - &	(0,2,3)	 & -	   & 	01110 &	- 	\\ \hline
(4,3,2,1) &	23 &	-  &	10111 &	 - &	(1,2,3)	 & -	   & 	11110 &	- 	\\ \hline
      
    \end{tabular}
    \label{tab:labeling}
  \end{center}
\end{table*}

\begin{table}[b]
  \begin{center}
    \caption{Examples of Gray labeling: (a) route $\mathbf{r} = $ (7, 5, 3, 6, 8, 1, 4, 2) and (b) route $\mathbf{r} = $ (5, 7, 3, 6, 8, 1, 4, 2).}
    \begin{tabular}{|c|c|c|c|} 
      \multicolumn{4}{l}{(a)Route $\mathbf{r} = $  ({\textbf{7, 5}}, 3, 6, 8, 1, 4, 2)} \\ \hline
      $i$ & $\mathcal{S}_i$         & $|\mathcal{S}_i|$ & $g_i(|\mathcal{S}_i|)$ \\ \hline
      1 & (Always $\varnothing$) & (Always 0) & (Always 0) \\ \hline
      2 & $\varnothing$          & 0 & 0 \\ \hline
      3 & \{1,2\}         & 2 & 11 \\ \hline
      4 & \{2\}           & 1 & 01 \\ \hline
      {\textbf{5}} & {\textbf{\{1,2,3,4\}}}     & {\textbf{4}}      & {\textbf{110}} \\ \hline
      6 & \{1,2,4\}       & 3 & 010 \\ \hline
      {\textbf{7}} & {\textbf{\{1,2,3,4,5,6\}}} & {\textbf{6}}      & {\textbf{101}} \\ \hline
      8 & \{1,2,4\}       & 3 & 010 \\ \hline
      
      \multicolumn{4}{c}{$\Downarrow$} \\ 
      \multicolumn{4}{c}{$l_G(\mathbf{r})=$ 01101{\textbf{110}}010{\textbf{101}}010}\\
      \multicolumn{4}{c}{}\\
    \end{tabular}

    \begin{tabular}{|c|c|c|c|} 
      \multicolumn{4}{l}{(b)Route $\mathbf{r} = $  ({\textbf{5, 7}}, 3, 6, 8, 1, 4, 2)} \\ \hline
      $i$ & $\mathcal{S}_i$         & $|\mathcal{S}_i|$ & $g_i(|\mathcal{S}_i|)$ \\ \hline
      1 & (Always $\varnothing$) & (Always 0) & (Always 0) \\ \hline
      2 & $\varnothing$          & 0 & 0 \\ \hline
      3 & \{1,2\}         & 2 & 11 \\ \hline
      4 & \{2\}           & 1 & 01 \\ \hline
      {\textbf{5}} & {\textbf{\{1,2,3,4\}}}     & {\textbf{4}}
      & {\textbf{110}} \\ \hline
      6 & \{1,2,4\}       & 3 & 010 \\ \hline
      {\textbf{7}} & {\textbf{\{1,2,3,4,6\}}} & {\textbf{5}}
      & {\textbf{111}} \\ \hline
      8 & \{1,2,4\}       & 3 & 010 \\ \hline    
      \multicolumn{4}{c}{$\Downarrow$} \\ 
      \multicolumn{4}{c}{$l_G(\mathbf{r})=$ 01101{\textbf{110}}010{\textbf{111}}010}

    \end{tabular}
    \label{tab:GL}    
  \end{center}
  
\end{table}

\subsection{Bit labelings in channel coding}
Bit labelings are essential in channel coding for spectrally efficient and reliable communications. While the logical layer treats bits, the channel requires symbols, where bits to symbols mapping rule is provided to make bit errors caused by a symbol error as less as possible. It is then better to provide similar labels with a small Hamming distance to neighboring symbols having a small Euclidean distance. A well-known method is binary (reflected) Gray coding \cite{Gray}, where $2^{\mathsf{m}}$-ary pulse amplitudes are labeled with $\mathsf{m}$ bits so that every Hamming distance between the nearest amplitudes is exactly 1. For example, amplitudes $\{3, 1, -1, -3\}$ are labeled as $\{00, 01, 10, 11 \}$ with natural coding and $\{00, 01, 11, 10 \}$ with Gray coding. This work extends the established concept of Gray coding to our binary labeling method, which is expected to be a key to avoid local solutions in optimization problems.

\subsection{Forward labeling}
Let $l_\mathrm{N}(\cdot)$ and $l_\mathrm{G}(\cdot)$ denote the bit labeling function obtained by applying natural labeling and Gray labeling, respectively. The output of these by inputting the route $\mathbf{r}$ provides the bit sequence $\mathbf{b}$. Tab.~\ref{tab:labeling} shows an example for $N=5$, including the forward labeling $\mathbf{r}\rightarrow\mathbf{b}$ and the inverse labeling $\mathbf{\underline{b}}\rightarrow\mathbf{r}$. Due to the definition, the bit sequence set is generally larger than that of the route set. Thus we employ $\mathbf{b}$ for the bit sequence having one-to-one correspondence to $\mathbf{r}$ (used in the forward labeling), and  $\mathbf{\underline{b}}$ for arbitrary combination of bits (used in the inverse labeling).

 Natural labeling directly corresponds $(N-1)!$ permutation cases in $N$-city TSP routes $\mathbf{r}$ to nonnegative integers $m \in \{ 0, 1, \ldots , (N-1)!-1 \}$, where $m$ is further described by the single bit sequence $\mathbf{b}$ with a length of $\ell_{\mathrm{N}}=\lceil \log (N-1)! \rceil (=\lceil \sum_{i=2}^{N-1}{\mathrm{log}i}\rceil)$, following the straight binary manner. The $\mathbf{b}$ is obtained by  $\mathbf{b}=n_{\ell_{\mathrm{N}}}(m)$, where $n_{\cdot}(\cdot)$ is the function obtaining a bit sequence having a length $\lambda$ from an arbitrary nonnegative integer $\gamma$, i.e.,
  \begin{equation}
    \begin{aligned}
      n_{\lambda}(\gamma) = \sigma_{0 \leq k < \lambda} (\eta_k(\gamma)).
    \end{aligned}
  \end{equation}     
The $\eta_k(\gamma)$ is the function to obtain the $k$-th bit from an arbitrary nonnegative integer $\gamma$ with the straight binary, i.e.,
  \begin{equation}
    \begin{aligned}
      \eta_{k}(\gamma) = \mathrm{mod} (\lfloor \gamma/2^k \rfloor, 2),
    \end{aligned}
  \end{equation}
where $\mathrm{mod}(\cdot,\cdot)$ denotes the modulo function. The $\sigma$ denotes the bit concatenation function from the most significant bit (the ($k_0-1$)-th bit) to the least significant bit (the 0-th bit), i.e.,
  \begin{equation}
    \begin{aligned}
      \sigma_{0 \leq k < k_0}(b_k) = b_{k_0 -1} b_{k_1 -2} \ldots b_{1} b_{0}
    \end{aligned}
  \end{equation}
with an arbitrary nonnegative integer $k_0$. The permutations are arranged in the lexicographical order, e.g., $l_\mathrm{N}((1, 2, 3, 4))=00000$, $l_\mathrm{N}((1, 2, 4, 3))=00001$, $l_\mathrm{N}((1, 3, 2, 4))=00010$, $\ldots$, $l_\mathrm{N}((4,3,2,1))=10111$ in the case of $N=5$. 

On the other hand, our proposal of Gray labeling combines the inversion number and Gray coding. The inversion number is the idea of discrete mathematics and relates to a kind of sort, the bubble sort, of sequences \cite{presortedness, bipartite}. Gray labeling mainly consists of the following two steps:\\

\noindent Step 1. For every $i$-th city ($i = 2, 3, \ldots, N-1$), enumerate the number of inversion cities, having an index $<i$ and visited after city $i$ except for the 0-th city. Configure the inversion city set $\mathcal{S}_i$ and quantify the set size $|\mathcal{S}_i|$.\\

\noindent Step 2. Convert each $|\mathcal{S}_i|$ to component bit sequence with the length $\lceil \log i \rceil$ by the Gray coding function $g_i(|\mathcal{S}_i|)$. Concatenate the component single bit sequence with the order from $i=2$ to $N-1$ to the single bit sequence having a length $\ell_\mathrm{G} = \sum_{i=2}^{N-1} \lceil \log i \rceil $. \\

This labeling method is explained with a small example; the city route $\mathbf{r}=(7, 5, 3, 6, 8, 1, 4, 2)$ for $N=9$ shown in Tab.~\ref{tab:GL}. Step 1 enumerates the inversion cities. For examples, there are 4 smaller numbers (3, 1, 4, 2) after 5, thus $\mathcal{S}_5=\{1, 2, 3, 4\}$ and $|\mathcal{S}_5|=4$, and no smaller numbers after 2, thus $\mathcal{S}_2=\varnothing$ and $|\mathcal{S}_2|=0$, where $\varnothing$ denotes the empty set. Enumerating every inversion number for $i=2$ to $N-1$ with the same manner, $|\mathcal{S}| = (|\mathcal{S}_2|, |\mathcal{S}_3|, \ldots, |\mathcal{S}_8|) = (0, 2, 1, 4, 3, 6, 3)$ is obtained. Step 2 converts $|\mathcal{S}_i|$ to the component bit sequence by the Gray coding function
  \begin{equation}
    \begin{aligned}
      g_i(|\mathcal{S}_i|) =  n_{\lambda}(|\mathcal{S}_i|) \oplus n_{\lambda}(\lfloor |\mathcal{S}_i| / 2 \rfloor), 
    \end{aligned}
  \end{equation}
where $\lambda = \lceil \log i \rceil$. $\oplus$ denotes the operator of bitwise exclusive OR. According to this definition, $g_2(|\mathcal{S}_2|) \rightarrow 0$, $g_3(|\mathcal{S}_3|) \rightarrow 11$, $\ldots$, $g_8(|\mathcal{S}_8|) \rightarrow 010$, where each length is bare minimum. Note that $i=1$ is ignored because $1$ has no inversion number. Finally, every obtained sequence for $i$ is concatenated from $i=2$ to $N-1=8$ into the single bit sequence $\mathbf{b}=$ $01101110010101010$ with the length $\ell_\mathrm{G} = \sum_{i=2}^{8} \lceil \log i \rceil = 17$. The conversion from $\mathbf{r} \rightarrow \mathbf{b}$ is injective but not surgective due to redundant description with binary variables.

\subsection{Inverse labeling}
Let the inverse labeling function of $l_\mathrm{N}(\mathbf{r}), l_\mathrm{G}(\mathbf{r})$ be $l_\mathrm{N}^{-1}(\mathbf{\underline{b}}), l_\mathrm{G}^{-1}(\mathbf{\underline{b}})$, to a given single bit sequence. When we employ annealing machines to optimize $\mathbf{\underline{b}}$, the obtained combination of binary variables can be arbitrary, i.e., there are totally $2^{\ell}$ cases with $\ell$ bits. 
The conversion from $\mathbf{\underline{b}} \rightarrow \mathbf{r}$ is surjective but not injective in general because possible cases with the concatenated single bit sequence can be more than the possible $(N-1)!$ routes. Thus we have to define the inverse function $\mathbf{\mathbf{b}} \rightarrow \mathbf{r}$ to be injective. We consider $\mathbf{b}$, its integer representation based on the straight binary manner $\underline{m}=n_{\ell_{\mathrm{N}}}^{-1}(\mathbf{\underline{b}})$, and let $m =  \mathrm{mod} (\underline{m}, (N-1)!)$ in natural labeling. Since $m < (N-1)!$, there exists a route $\mathbf{r}=l_{\mathrm{N}}^{-1}(\mathbf{b})$, where $\mathbf{b}=n_{\ell_{\mathrm{N}}}(m)$. In Gray labeling, the inverse operation recovers the route $\mathbf{r}=g_i^{-1}(|\underline{\mathcal{S}_i}|)$, where $|\underline{\mathcal{S}_i}| = \mathrm{mod} (|\mathcal{S}_i|, i)$. An example is again referred to Tab.~\ref{tab:labeling}, e.g., $\mathbf{\underline{b}}=11011$ corresponds to $\underline{m}=27$.  In natural labeling, $m=\mathrm{mod} (27, (5-1)!) = 3$, and $l_\mathrm{N}^{-1}(11011) = l_\mathrm{N}^{-1}(00011) = (1, 3, 4, 2)$. In Gray labeling, $|\mathcal{S}| = (|\mathcal{S}_2|, |\mathcal{S}_3|, |\mathcal{S}_4|) = (1, 3, 2)$, and $|\mathcal{\underline{S}}| = (|\underline{\mathcal{S}_2}|, |\underline{\mathcal{S}_3}|, |\underline{\mathcal{S}_4}|) = (1, 0, 2)$. Therefore, $l_\mathrm{G}^{-1}(11011)= l_\mathrm{G}^{-1}(10011) = (2, 4, 1, 3)$.

The bit length for Gray labeling $\ell_\mathrm{G}=\sum_{i=2}^{N-1}\lceil \log i \rceil$ is greater than or equal to that for natural labeling, $\ell_\mathrm{N}= \lceil \log (N-1)! \rceil (=\lceil \sum_{i=2}^{N-1}{\mathrm{log}i}\rceil)$. These lengths $\ell_\mathrm{N}$ and $\ell_\mathrm{G}$ are approximated to $O(\mathrm{log}(N!)) = O(N\mathrm{log}N)$. The proposed method of combining the inversion number and Gray coding is originated from the idea: \emph{similar bits for similar routes}. A pair of similar routes just in the relationship of swapping two cities consecutively visited, the Hamming distance between their bit sequence equals exactly 1 for the proposed Gray labeling. 
Let $r_j$ and $r_{j+1}$ be the indices of a pair of cities consecutively visited. In Gray labeling, $|\mathcal{S}_{r_{j+1}}|$ under $r_j < r_{j+1}$ is smaller by 1 than $|\mathcal{S}_{r_{j+1}}|$ under $r_j > r_{j+1}$, and the other $|\mathcal{S}_i|$ maintains. When the resultant bit sequence pair obtained from a difference in $|\mathcal{S}_{r_{j+1}}|$, the Hamming distance between those is guaranteed to be 1 with Gray coding and not guaranteed with natural coding. 
Tab.~\ref{tab:GL} shows an example of similar routes (a) $\mathbf{r}=(7, 5, 3, 6, 8, 1, 4, 2)$ and (b) $\mathbf{r}=(5, 7, 3, 6, 8, 1, 4, 2)$. In this case, only $|\mathcal{S}_7|$ is different from each other and the other $|\mathcal{S}_i|$ are identical, and the Hamming distance between their concatenated bit sequence is exactly 1.

\section{Local solution metric and analysis} \label{sec:LS}
Performance of an optimization solver is characterized by the balance of the solution quality and the required computational resource, which can be translated into and the Ising energy and the number of iterations for a solver based on an annealing machine. Our proposed Gray labeling in the previous section would be useful, especially for avoiding local solutions. This section introduces the \textit{local solution metric} to quantify the expected performance without running actual optimization procedure.

Our local solution metric is given by the number of local solutions normalized by the number of all solution candidates, which will be explained later. A solution is defined as a local solution if all of the nearest solutions (with the Hamming distance of 1 from the solution under examination) have worse or equal solution quality. 
Instead of $d(\mathbf{r})$, let $d(\mathbf{\underline{b}})$ simply denote the total traveling distance in each route $\mathbf{r} = l_{N}^{-1}(\mathbf{\underline{b}})$ for natural labeling or $\mathbf{r} = l_{G}^{-1}(\mathbf{\underline{b}})$ for Gray labeling according to Eq.~\eqref{eq:dist}. 
The local solution flag is defined as
  \begin{equation}
    \begin{aligned}
      f(\mathbf{\underline{b}}) = \prod_{k=0}^{\ell-1}{\delta (d(\mathbf{\underline{b}}) \le d(\mathbf{\underline{b}} \oplus 2^k))}
    \end{aligned}
  \end{equation}
for the single bit sequence $\mathbf{b}$ and its length $\ell$ ($\ell_{\mathrm{N}}$ for natural and $\ell_{\mathrm{G}}$ for Gray labeling, respectively), where $\delta (·)$ is 1 if the argument is true and 0 otherwise. The $\oplus 2^k$ flips the $k$-th bit only, to obtain similar single bit sequence apart by the Hamming distance of 1 from $\mathbf{b}$.

An example to compute $f$ for $N=5$ is explained below. 
When we treat $\mathbf{\underline{b}}=00110$, the corresponding route $\mathbf{r}$ is (2, 1, 3, 4) in natural labeling, and (4, 1, 3, 2) in Gray labeling, respectively. The set of $\mathbf{\underline{b}} ^\prime =\mathbf{\underline{b}} \oplus 2^k$ is \{00111, 00100, 00010, 01110, 10110\}, and the set of $\mathbf{r}$ is thus \{(2, 1, 4, 3), (1, 4, 2, 3), (1, 3, 2, 4), (3, 2, 1, 4), (4, 3, 1, 2)\} given by $l_{\mathrm{N}}^{-1}(\mathbf{\underline{b}}^{\prime})$ for natural labeling and \{(1, 4, 3, 2), (1, 3, 2, 4), (4, 1, 2, 3), (4, 3, 1, 2), (4, 2, 3, 1)\} given by $l_{\mathrm{G}}^{-1}(\mathbf{\underline{b}}^{\prime})$ for Gray labeling, respectively. 
An arbitrary similar route $\mathbf{r}^{\prime}$ with the reference route $\mathbf{r}$ is given by swapping a pair of cities consecutively visited. Here in Gray labeling, any bit sequence from $\mathbf{r}^{\prime}$ is described by either one of $\mathbf{\underline{b}}^{\prime}$, corresponding to the Hamming distance between $\mathbf{\underline{b}}$ and $\mathbf{\underline{b}}^{\prime}$ equals exactly 1. This feature is unique to Gray labeling.

\begin{figure}[t]
  \begin{center}
    \includegraphics[clip, width=8.4cm]{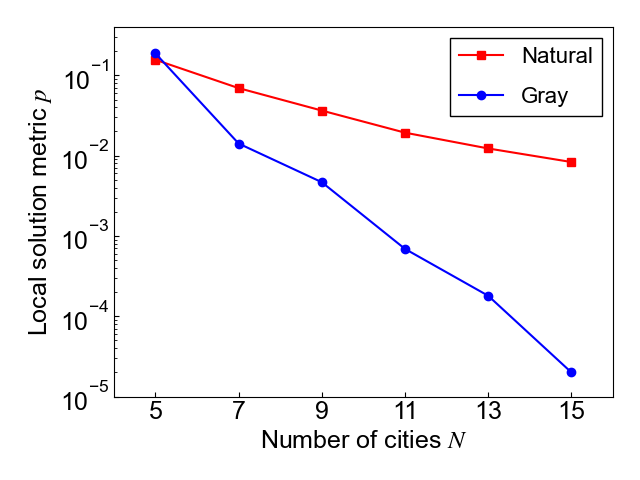}
    \caption{The local solution metric $p$ for two labeling methods as a function of the number of cities $N$.}
    \label{fig:local_minima}
  \end{center}
\end{figure}

Based on $f$, the local solution metric $p$ is given by
  \begin{equation}
    \begin{aligned}
      p = \mathbb{E}_{\mathbf{b}} \left[ f(\mathbf{\underline{b}}) \right] ,
    \end{aligned}
  \end{equation}
  where $\mathbb{E} \left[ \cdot \right]$ denotes the expectation.
  Fig. \ref{fig:local_minima} shows the metric $p$ in each labeling for $N=5$ to $15$. There are too many cases to quantify full cases for an $N \ge 11$, so we sampled at maximum $10^5$ cases randomly. The metric $p$ decreases as increasing $N$, where Gray labeling shows more rapid decrease than natural labeling. This feature would be advantageous in better convergence in optimization because of avoiding local solutions when exploring ones through bit flips with an annealing machine. Note that, under the condition of a small number of cities, swapping the cities consecutively visited results in a significant change in the route and the distance, e.g., the number of local solutions is 5 for natural labeling and 6 for Gray labeling for $N=5$, in all of the $2^5$ cases.

\section{Numerical simulations} \label{sec:NS}

\begin{figure*}[t]
  \begin{center}   

    \includegraphics[clip, width=14cm]{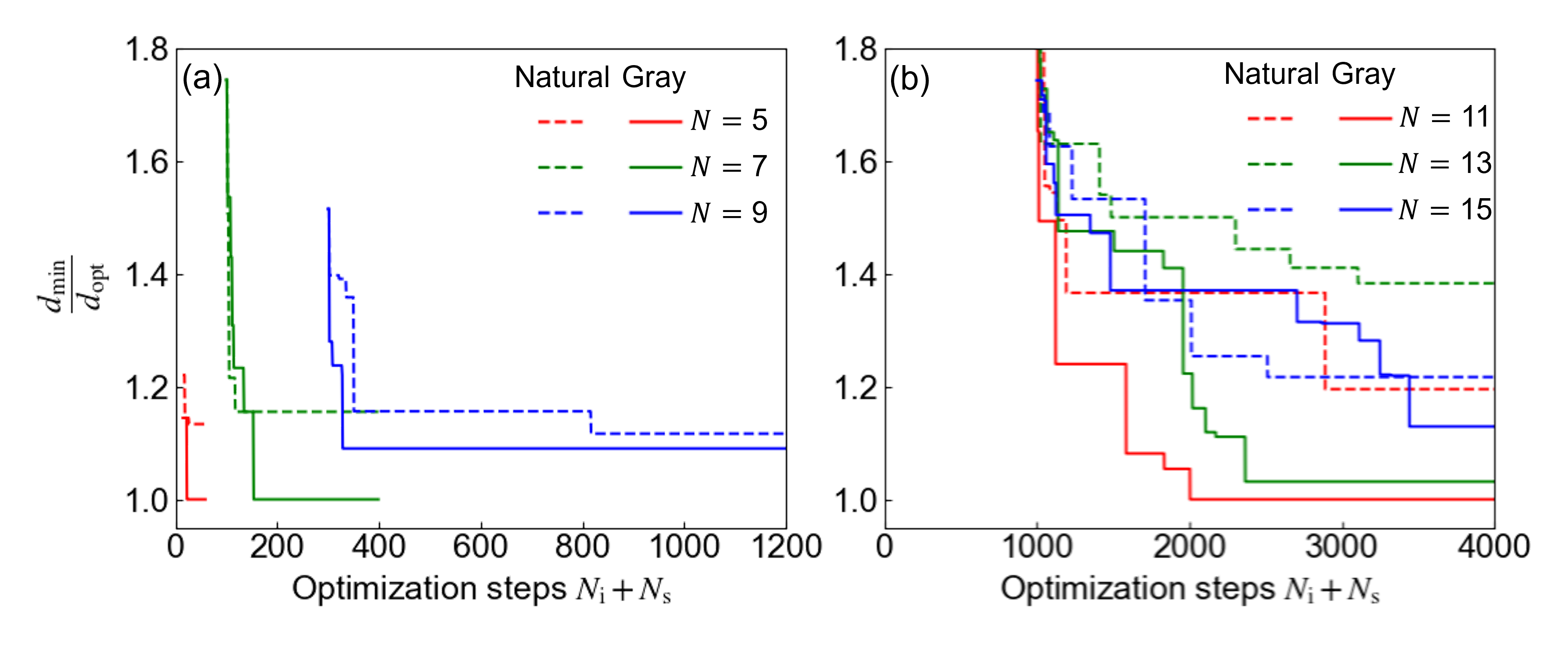}
    \caption{Numerically obtained shortest distance $d_{\mathrm{min}}$ until the step normalized by the optimum one in FMA-based TSP for (a) $N = $ 5 to 9 and (b) $N = $ 11 to 15. }
    \label{fig:Result}

    \includegraphics[clip, width=14cm]{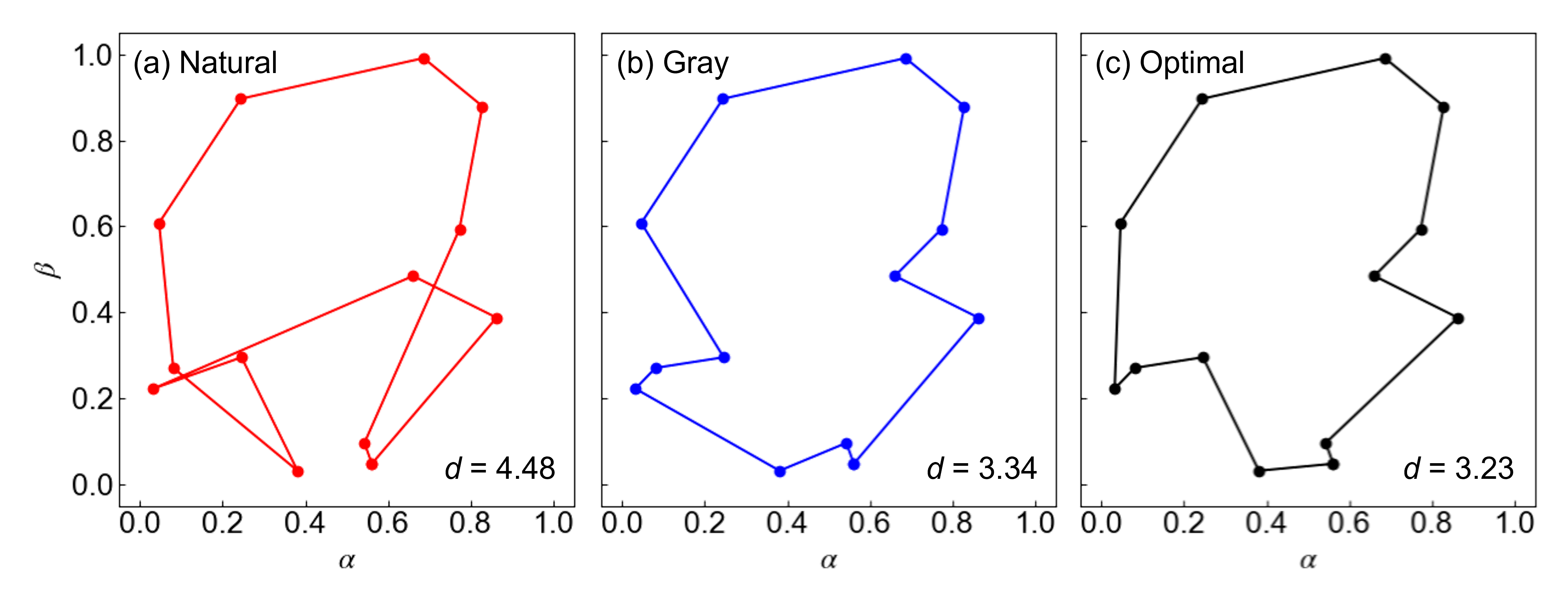}
    \caption{Numerically obtained results of TSP routes in 13 cities: routes obtained by (a) natural labeling and (b) Gray labeling compared with (c) the optimal route, respectively. }
    \label{fig:route}
  \end{center}
\end{figure*}

This section numerically compares natural labeling and Gray labeling in terms of the obtained solution quality and the convergence speed with FMA. 
As shown in Section~\ref{sec:FMA}, a solution candidates of the single bit sequence $\mathbf{b}$ were randomly generated and used in bits $\mathbf{x}$ for the initial training. After the training, an acquisition function $y$ was constructed with the FM. Subsequently, the bit sequence $\mathbf{b}$ minimizing the acquisition function $y$ was estimated using an annealing machine, and the route-distance pair was added to the training data. 
The number of data points for the initial training and that for the solution search are denoted as $N_\mathrm{i}$ and $N_\mathrm{s}$, respectively. These parameters were set to ($N$, $N_\mathrm{i}$, $N_\mathrm{s}$) = (5, 15, 45), (7, 100, 300), (9, 300, 900), (11, 1000, 3000), (13, 1000, 3000), (15, 1000, 3000).

The comparison results between the two labeling methods are shown in Fig. \ref{fig:Result}. Here, $d_\mathrm{opt}$ and $d_\mathrm{min}$ indicate the globally optimum distance and the minimum distance obtained until the step, respectively. Gray labeling shows mostly smaller $d_{\mathrm{min}}$ or faster convergence than natural labeling in any optimization steps for every $N$. Especially, Gray labeling reached the global optimal solutions for $N=5, 7, 11$, while natural labeling did not. For the trial of $N=15$, natural labeling and Gray labeling show almost the same balance of the solution quality and the convergence speed. 
Fig. \ref{fig:route} shows the finally obtained routes by (a) natural and (b) Gray labeling at the final optimization step in our trial, and (c) the globally optimal route for $N = 13$. The corresponding distances $d$ were 4.48, 3.34, and 3.23, respectively. Compared with the optimal route, natural labeling and Gray labeling yielded longer routes by 39\% and 3\%, respectively.
Overall, Gray labeling is expected to avoid local solutions more frequently than natural labeling, resulting in the better quality–speed balance, as predicted from the local solution metric according to the previous section.

\section{Conclusion} \label{sec:conc}
This work addresses the local solution characterization and its avoidance by bit labeling method in FMA, a QUBO solver combined with machine learning. Especially we focused on TSP, where FMA could reduce the required number of bits from $N^2$ to $N\mathrm{log}N$ for $N$-city TSP. Within the context of the FMA-based TSP, two labeling method of natural and Gray labeling were compared. While natural labeling converted $(N-1)!$ routes to the lexicographical integer and straight-binary label, Gray labeling employed inversion number and Gray coding to realize the idea of \emph{similar bits for similar routes} with the help of slightly larger number of bits. The originally introduced metric simply quantified the local solution ratio without performing actual optimization, where Gray labeling showed rapid reduction of the ratio compared with natural labeling as increasing $N$. Through the actual numerical optimization, Gray labeling showed often better balance of the solution quality and the convergence speed because of the feature of fewer probability of being captured at local solutions. Our results suggest that both the proposed Gray labeling and the proposed metric are useful for QUBO solvers combined with machine learning such as FMA. \\

\begin{acknowledgements}

The authors thank Mr. Koichi Yanagisawa, Mr. Isamu Kudo, and Dr. Narumitsu Ikeda of Mitsubishi Electric Corp. for the fruitiful discussion.

\end{acknowledgements}

%%%%%%%%%% If using BibTeX:
\bibliographystyle{apsrev}
\bibliography{references}

\begin{thebibliography}{36}
\expandafter\ifx\csname natexlab\endcsname\relax\def\natexlab#1{#1}\fi
\expandafter\ifx\csname bibnamefont\endcsname\relax
  \def\bibnamefont#1{#1}\fi
\expandafter\ifx\csname bibfnamefont\endcsname\relax
  \def\bibfnamefont#1{#1}\fi
\expandafter\ifx\csname citenamefont\endcsname\relax
  \def\citenamefont#1{#1}\fi
\expandafter\ifx\csname url\endcsname\relax
  \def\url#1{\texttt{#1}}\fi
\expandafter\ifx\csname urlprefix\endcsname\relax\def\urlprefix{URL }\fi
\providecommand{\bibinfo}[2]{#2}
\providecommand{\eprint}[2][]{\url{#2}}

\bibitem[{\citenamefont{Yu et~al.}(2019)\citenamefont{Yu, Yu, and
  Gu}}]{transportation}
\bibinfo{author}{\bibfnamefont{J.~J.~Q.} \bibnamefont{Yu}},
  \bibinfo{author}{\bibfnamefont{W.}~\bibnamefont{Yu}}, \bibnamefont{and}
  \bibinfo{author}{\bibfnamefont{J.}~\bibnamefont{Gu}}, \bibinfo{journal}{IEEE
  Transactions on Intelligent Transportation Systems}
  \textbf{\bibinfo{volume}{20}}, \bibinfo{pages}{3806} (\bibinfo{year}{2019}).

\bibitem[{\citenamefont{Tao et~al.}(2013)\citenamefont{Tao, LaiLi, Xu, and
  Zhang}}]{manufacturing}
\bibinfo{author}{\bibfnamefont{F.}~\bibnamefont{Tao}},
  \bibinfo{author}{\bibfnamefont{Y.}~\bibnamefont{LaiLi}},
  \bibinfo{author}{\bibfnamefont{L.}~\bibnamefont{Xu}}, \bibnamefont{and}
  \bibinfo{author}{\bibfnamefont{L.}~\bibnamefont{Zhang}},
  \bibinfo{journal}{IEEE Transactions on Industrial Informatics}
  \textbf{\bibinfo{volume}{9}}, \bibinfo{pages}{2023} (\bibinfo{year}{2013}).

\bibitem[{\citenamefont{Woeginger}(2003)}]{NP_survey}
\bibinfo{author}{\bibfnamefont{G.~J.} \bibnamefont{Woeginger}},
  \emph{\bibinfo{title}{Exact Algorithms for NP-Hard Problems: A Survey}}
  (\bibinfo{publisher}{Springer Berlin Heidelberg}, \bibinfo{year}{2003}).

\bibitem[{\citenamefont{Alexandersson}(2020)}]{NP_graph}
\bibinfo{author}{\bibfnamefont{P.}~\bibnamefont{Alexandersson}},
  \bibinfo{journal}{arXiv:2001.04120}  (\bibinfo{year}{2020}).

\bibitem[{\citenamefont{Sanyal and Roy}(2022)}]{neuro_ising}
\bibinfo{author}{\bibfnamefont{S.}~\bibnamefont{Sanyal}} \bibnamefont{and}
  \bibinfo{author}{\bibfnamefont{K.}~\bibnamefont{Roy}}, \bibinfo{journal}{IEEE
  Transactions on Computer-Aided Design of Integrated Circuits and Systems}
  \textbf{\bibinfo{volume}{41}}, \bibinfo{pages}{5408} (\bibinfo{year}{2022}).

\bibitem[{\citenamefont{Zhang and Han}(2022)}]{SB_TSP}
\bibinfo{author}{\bibfnamefont{T.}~\bibnamefont{Zhang}} \bibnamefont{and}
  \bibinfo{author}{\bibfnamefont{J.}~\bibnamefont{Han}},
  \bibinfo{journal}{Design, Automation and Test in Europe Conference} p.
  \bibinfo{pages}{548} (\bibinfo{year}{2022}).

\bibitem[{\citenamefont{Kitai et~al.}(2020)\citenamefont{Kitai, Guo, Ju,
  Tanaka, Tsuda, Shiomi, and Tamura}}]{FMA_META}
\bibinfo{author}{\bibfnamefont{K.}~\bibnamefont{Kitai}},
  \bibinfo{author}{\bibfnamefont{J.}~\bibnamefont{Guo}},
  \bibinfo{author}{\bibfnamefont{S.}~\bibnamefont{Ju}},
  \bibinfo{author}{\bibfnamefont{S.}~\bibnamefont{Tanaka}},
  \bibinfo{author}{\bibfnamefont{K.}~\bibnamefont{Tsuda}},
  \bibinfo{author}{\bibfnamefont{J.}~\bibnamefont{Shiomi}}, \bibnamefont{and}
  \bibinfo{author}{\bibfnamefont{R.}~\bibnamefont{Tamura}},
  \bibinfo{journal}{Physical Review Research} \textbf{\bibinfo{volume}{2}},
  \bibinfo{pages}{013319} (\bibinfo{year}{2020}).

\bibitem[{\citenamefont{Seki et~al.}(2022)\citenamefont{Seki, Tamura, and
  Tanaka}}]{FMA_ENC}
\bibinfo{author}{\bibfnamefont{Y.}~\bibnamefont{Seki}},
  \bibinfo{author}{\bibfnamefont{R.}~\bibnamefont{Tamura}}, \bibnamefont{and}
  \bibinfo{author}{\bibfnamefont{S.}~\bibnamefont{Tanaka}},
  \bibinfo{journal}{arXiv:2209.01016}  (\bibinfo{year}{2022}).

\bibitem[{\citenamefont{Inoue et~al.}(2022)\citenamefont{Inoue, Seki, Tanaka,
  Togawa, Ishizaki, and Noda}}]{FMA_PHOTONIC}
\bibinfo{author}{\bibfnamefont{T.}~\bibnamefont{Inoue}},
  \bibinfo{author}{\bibfnamefont{Y.}~\bibnamefont{Seki}},
  \bibinfo{author}{\bibfnamefont{S.}~\bibnamefont{Tanaka}},
  \bibinfo{author}{\bibfnamefont{N.}~\bibnamefont{Togawa}},
  \bibinfo{author}{\bibfnamefont{K.}~\bibnamefont{Ishizaki}}, \bibnamefont{and}
  \bibinfo{author}{\bibfnamefont{S.}~\bibnamefont{Noda}},
  \bibinfo{journal}{Optics Express} \textbf{\bibinfo{volume}{30}},
  \bibinfo{pages}{43503} (\bibinfo{year}{2022}).

\bibitem[{\citenamefont{Kadowaki and Ambai}(2022)}]{bbo_kadowaki}
\bibinfo{author}{\bibfnamefont{T.}~\bibnamefont{Kadowaki}} \bibnamefont{and}
  \bibinfo{author}{\bibfnamefont{M.}~\bibnamefont{Ambai}},
  \bibinfo{journal}{Scientific Reports} \textbf{\bibinfo{volume}{12}},
  \bibinfo{pages}{15482} (\bibinfo{year}{2022}).

\bibitem[{\citenamefont{Matsumori et~al.}(2022)\citenamefont{Matsumori, Taki,
  and Kadowaki}}]{bbo_matsumori}
\bibinfo{author}{\bibfnamefont{T.}~\bibnamefont{Matsumori}},
  \bibinfo{author}{\bibfnamefont{M.}~\bibnamefont{Taki}}, \bibnamefont{and}
  \bibinfo{author}{\bibfnamefont{T.}~\bibnamefont{Kadowaki}},
  \bibinfo{journal}{Scientific Reports} \textbf{\bibinfo{volume}{12}},
  \bibinfo{pages}{12143} (\bibinfo{year}{2022}).

\bibitem[{\citenamefont{Nawa et~al.}(2023)\citenamefont{Nawa, Suzuki, Masuda,
  Tanaka, and Miura}}]{FMA_MTJ}
\bibinfo{author}{\bibfnamefont{K.}~\bibnamefont{Nawa}},
  \bibinfo{author}{\bibfnamefont{T.}~\bibnamefont{Suzuki}},
  \bibinfo{author}{\bibfnamefont{K.}~\bibnamefont{Masuda}},
  \bibinfo{author}{\bibfnamefont{S.}~\bibnamefont{Tanaka}}, \bibnamefont{and}
  \bibinfo{author}{\bibfnamefont{Y.}~\bibnamefont{Miura}},
  \bibinfo{journal}{Physical Review Applied} \textbf{\bibinfo{volume}{20}},
  \bibinfo{pages}{024044} (\bibinfo{year}{2023}).

\bibitem[{\citenamefont{Ramani et~al.}(2008)\citenamefont{Ramani, Blu, and
  Unser}}]{BBO}
\bibinfo{author}{\bibfnamefont{S.}~\bibnamefont{Ramani}},
  \bibinfo{author}{\bibfnamefont{T.}~\bibnamefont{Blu}}, \bibnamefont{and}
  \bibinfo{author}{\bibfnamefont{M.}~\bibnamefont{Unser}},
  \bibinfo{journal}{IEEE Transactions on Image Processing}
  \textbf{\bibinfo{volume}{17}}, \bibinfo{pages}{1540} (\bibinfo{year}{2008}).

\bibitem[{\citenamefont{Terayama et~al.}(2021)\citenamefont{Terayama, Sumita,
  Tamura, and Tsuda}}]{BBO_FMQA_REVIEW}
\bibinfo{author}{\bibfnamefont{K.}~\bibnamefont{Terayama}},
  \bibinfo{author}{\bibfnamefont{M.}~\bibnamefont{Sumita}},
  \bibinfo{author}{\bibfnamefont{R.}~\bibnamefont{Tamura}}, \bibnamefont{and}
  \bibinfo{author}{\bibfnamefont{K.}~\bibnamefont{Tsuda}},
  \bibinfo{journal}{Accounts of Chemical Research}
  \textbf{\bibinfo{volume}{54}}, \bibinfo{pages}{1334} (\bibinfo{year}{2021}).

\bibitem[{\citenamefont{Doi et~al.}(2023)\citenamefont{Doi, Nakao, Tanaka,
  Sako, and Ohzeki}}]{bbo_doi}
\bibinfo{author}{\bibfnamefont{M.}~\bibnamefont{Doi}},
  \bibinfo{author}{\bibfnamefont{Y.}~\bibnamefont{Nakao}},
  \bibinfo{author}{\bibfnamefont{T.}~\bibnamefont{Tanaka}},
  \bibinfo{author}{\bibfnamefont{M.}~\bibnamefont{Sako}}, \bibnamefont{and}
  \bibinfo{author}{\bibfnamefont{M.}~\bibnamefont{Ohzeki}},
  \bibinfo{journal}{Frontiers in Computer Science} \textbf{\bibinfo{volume}{5}}
  (\bibinfo{year}{2023}).

\bibitem[{\citenamefont{N{\"u}{\ss}lein
  et~al.}(2023)\citenamefont{N{\"u}{\ss}lein, Roch, Gabor, Stein,
  Linnhoff-Popien, and Feld}}]{bbo_cross_entropy}
\bibinfo{author}{\bibfnamefont{J.}~\bibnamefont{N{\"u}{\ss}lein}},
  \bibinfo{author}{\bibfnamefont{C.}~\bibnamefont{Roch}},
  \bibinfo{author}{\bibfnamefont{T.}~\bibnamefont{Gabor}},
  \bibinfo{author}{\bibfnamefont{J.}~\bibnamefont{Stein}},
  \bibinfo{author}{\bibfnamefont{C.}~\bibnamefont{Linnhoff-Popien}},
  \bibnamefont{and} \bibinfo{author}{\bibfnamefont{S.}~\bibnamefont{Feld}}, in
  \emph{\bibinfo{booktitle}{Computational Science -- ICCS 2023}}, edited by
  \bibinfo{editor}{\bibfnamefont{J.}~\bibnamefont{Miky{\v{s}}ka}},
  \bibinfo{editor}{\bibfnamefont{C.}~\bibnamefont{de~Mulatier}},
  \bibinfo{editor}{\bibfnamefont{M.}~\bibnamefont{Paszynski}},
  \bibinfo{editor}{\bibfnamefont{V.~V.} \bibnamefont{Krzhizhanovskaya}},
  \bibinfo{editor}{\bibfnamefont{J.~J.} \bibnamefont{Dongarra}},
  \bibnamefont{and} \bibinfo{editor}{\bibfnamefont{P.~M.} \bibnamefont{Sloot}}
  (\bibinfo{publisher}{Springer Nature Switzerland}, \bibinfo{address}{Cham},
  \bibinfo{year}{2023}), pp. \bibinfo{pages}{48--55}.

\bibitem[{\citenamefont{Izawa et~al.}(2022)\citenamefont{Izawa, Kitai, Tanaka,
  Tamura, and Tsuda}}]{bbo_izawa}
\bibinfo{author}{\bibfnamefont{S.}~\bibnamefont{Izawa}},
  \bibinfo{author}{\bibfnamefont{K.}~\bibnamefont{Kitai}},
  \bibinfo{author}{\bibfnamefont{S.}~\bibnamefont{Tanaka}},
  \bibinfo{author}{\bibfnamefont{R.}~\bibnamefont{Tamura}}, \bibnamefont{and}
  \bibinfo{author}{\bibfnamefont{K.}~\bibnamefont{Tsuda}},
  \bibinfo{journal}{Physical Review Research} \textbf{\bibinfo{volume}{4}},
  \bibinfo{pages}{023062} (\bibinfo{year}{2022}).

\bibitem[{\citenamefont{Rendle}(2010)}]{FM}
\bibinfo{author}{\bibfnamefont{S.}~\bibnamefont{Rendle}}, in
  \emph{\bibinfo{booktitle}{Proceedings of IEEE International Conference on
  Data Mining}} (\bibinfo{publisher}{IEEE}, \bibinfo{year}{2010}), pp.
  \bibinfo{pages}{995--1000}.

\bibitem[{\citenamefont{Lucas}(2014)}]{lucas}
\bibinfo{author}{\bibfnamefont{A.}~\bibnamefont{Lucas}},
  \bibinfo{journal}{Frontiers in Physics} \textbf{\bibinfo{volume}{2}}
  (\bibinfo{year}{2014}).

\bibitem[{\citenamefont{Tan et~al.}(2021)\citenamefont{Tan, Lemonde, Thanasilp,
  Tangpanitanon, and Angelakis}}]{QUBIT_ENCODING}
\bibinfo{author}{\bibfnamefont{B.}~\bibnamefont{Tan}},
  \bibinfo{author}{\bibfnamefont{M.-A.} \bibnamefont{Lemonde}},
  \bibinfo{author}{\bibfnamefont{S.}~\bibnamefont{Thanasilp}},
  \bibinfo{author}{\bibfnamefont{J.}~\bibnamefont{Tangpanitanon}},
  \bibnamefont{and} \bibinfo{author}{\bibfnamefont{D.~G.}
  \bibnamefont{Angelakis}}, \bibinfo{journal}{Quantum}
  \textbf{\bibinfo{volume}{5}}, \bibinfo{pages}{454} (\bibinfo{year}{2021}).

\bibitem[{\citenamefont{Schnaus
  et~al.}(2024{\natexlab{a}})\citenamefont{Schnaus, Palackal, Poggel, Runge,
  Ehm, Lorenz, and Mendl}}]{TSP_VQE}
\bibinfo{author}{\bibfnamefont{M.}~\bibnamefont{Schnaus}},
  \bibinfo{author}{\bibfnamefont{L.}~\bibnamefont{Palackal}},
  \bibinfo{author}{\bibfnamefont{B.}~\bibnamefont{Poggel}},
  \bibinfo{author}{\bibfnamefont{X.}~\bibnamefont{Runge}},
  \bibinfo{author}{\bibfnamefont{H.}~\bibnamefont{Ehm}},
  \bibinfo{author}{\bibfnamefont{J.~M.} \bibnamefont{Lorenz}},
  \bibnamefont{and} \bibinfo{author}{\bibfnamefont{C.~B.} \bibnamefont{Mendl}},
  \bibinfo{journal}{arXiv:2404.05448}  (\bibinfo{year}{2024}{\natexlab{a}}).

\bibitem[{\citenamefont{Kikuchi et~al.}(2023)\citenamefont{Kikuchi, Togawa, and
  Tanaka}}]{kikuchi}
\bibinfo{author}{\bibfnamefont{S.}~\bibnamefont{Kikuchi}},
  \bibinfo{author}{\bibfnamefont{N.}~\bibnamefont{Togawa}}, \bibnamefont{and}
  \bibinfo{author}{\bibfnamefont{S.}~\bibnamefont{Tanaka}},
  \bibinfo{journal}{arXiv:2304.12796}  (\bibinfo{year}{2023}).

\bibitem[{\citenamefont{Applegate et~al.}(2007)\citenamefont{Applegate, Bixby,
  Chvátal, and Cook}}]{applegate}
\bibinfo{author}{\bibfnamefont{D.~L.} \bibnamefont{Applegate}},
  \bibinfo{author}{\bibfnamefont{R.~E.} \bibnamefont{Bixby}},
  \bibinfo{author}{\bibfnamefont{V.}~\bibnamefont{Chvátal}}, \bibnamefont{and}
  \bibinfo{author}{\bibfnamefont{W.~J.} \bibnamefont{Cook}},
  \emph{\bibinfo{title}{The Traveling Salesman Problem: A Computational Study}}
  (\bibinfo{publisher}{Princeton University Press}, \bibinfo{year}{2007}).

\bibitem[{\citenamefont{Laporte}(1992)}]{laporte}
\bibinfo{author}{\bibfnamefont{G.}~\bibnamefont{Laporte}},
  \bibinfo{journal}{European Journal of Operational Research}
  \textbf{\bibinfo{volume}{59}}, \bibinfo{pages}{231} (\bibinfo{year}{1992}).

\bibitem[{\citenamefont{Held and Karp}(1962)}]{held_karp}
\bibinfo{author}{\bibfnamefont{M.}~\bibnamefont{Held}} \bibnamefont{and}
  \bibinfo{author}{\bibfnamefont{R.~M.} \bibnamefont{Karp}},
  \bibinfo{journal}{Journal of the Society for Industrial and Applied
  Mathematics} \textbf{\bibinfo{volume}{10}}, \bibinfo{pages}{196}
  (\bibinfo{year}{1962}).

\bibitem[{\citenamefont{Bellman}(1962)}]{bellman}
\bibinfo{author}{\bibfnamefont{R.}~\bibnamefont{Bellman}},
  \bibinfo{journal}{Journal of the ACM (JACM)} \textbf{\bibinfo{volume}{9}},
  \bibinfo{pages}{61} (\bibinfo{year}{1962}).

\bibitem[{\citenamefont{Rosenkrantz et~al.}(1977)\citenamefont{Rosenkrantz,
  Stearns, and Lewis}}]{TSP_greedy}
\bibinfo{author}{\bibfnamefont{D.~J.} \bibnamefont{Rosenkrantz}},
  \bibinfo{author}{\bibfnamefont{R.~E.} \bibnamefont{Stearns}},
  \bibnamefont{and} \bibinfo{author}{\bibfnamefont{P.~M.} \bibnamefont{Lewis}},
  \bibinfo{journal}{SIAM journal on computing} \textbf{\bibinfo{volume}{6}},
  \bibinfo{pages}{563} (\bibinfo{year}{1977}).

\bibitem[{\citenamefont{Lin and Kernighan}(1973)}]{TSP_local}
\bibinfo{author}{\bibfnamefont{S.}~\bibnamefont{Lin}} \bibnamefont{and}
  \bibinfo{author}{\bibfnamefont{B.~W.} \bibnamefont{Kernighan}},
  \bibinfo{journal}{Operations research} \textbf{\bibinfo{volume}{21}},
  \bibinfo{pages}{498} (\bibinfo{year}{1973}).

\bibitem[{\citenamefont{Grefenstette et~al.}(1985)\citenamefont{Grefenstette,
  Gopal, Rosmaita, and Van~Gucht}}]{TSP_GA}
\bibinfo{author}{\bibfnamefont{J.~J.} \bibnamefont{Grefenstette}},
  \bibinfo{author}{\bibfnamefont{R.}~\bibnamefont{Gopal}},
  \bibinfo{author}{\bibfnamefont{B.~J.} \bibnamefont{Rosmaita}},
  \bibnamefont{and}
  \bibinfo{author}{\bibfnamefont{D.}~\bibnamefont{Van~Gucht}}, in
  \emph{\bibinfo{booktitle}{Proceedings of the first International Conference
  on Genetic Algorithms and their Applications}} (\bibinfo{year}{1985}), pp.
  \bibinfo{pages}{160--168}.

\bibitem[{\citenamefont{Dorigo and Gambardella}(1997)}]{TSP_ant}
\bibinfo{author}{\bibfnamefont{M.}~\bibnamefont{Dorigo}} \bibnamefont{and}
  \bibinfo{author}{\bibfnamefont{L.~M.} \bibnamefont{Gambardella}},
  \bibinfo{journal}{IEEE Transactions on evolutionary computation}
  \textbf{\bibinfo{volume}{1}}, \bibinfo{pages}{53} (\bibinfo{year}{1997}).

\bibitem[{\citenamefont{Martoňák et~al.}(2004)\citenamefont{Martoňák,
  Santoro, and Tosatti}}]{TSP_QA}
\bibinfo{author}{\bibfnamefont{R.}~\bibnamefont{Martoňák}},
  \bibinfo{author}{\bibfnamefont{G.~E.} \bibnamefont{Santoro}},
  \bibnamefont{and} \bibinfo{author}{\bibfnamefont{E.}~\bibnamefont{Tosatti}},
  \bibinfo{journal}{Physical Review E} \textbf{\bibinfo{volume}{70}},
  \bibinfo{pages}{057701} (\bibinfo{year}{2004}).

\bibitem[{\citenamefont{Ramezani et~al.}(2024)\citenamefont{Ramezani, Salami,
  Shokhmkar, Moradi, and Bahrampour}}]{ramezani}
\bibinfo{author}{\bibfnamefont{M.}~\bibnamefont{Ramezani}},
  \bibinfo{author}{\bibfnamefont{S.}~\bibnamefont{Salami}},
  \bibinfo{author}{\bibfnamefont{M.}~\bibnamefont{Shokhmkar}},
  \bibinfo{author}{\bibfnamefont{M.}~\bibnamefont{Moradi}}, \bibnamefont{and}
  \bibinfo{author}{\bibfnamefont{A.}~\bibnamefont{Bahrampour}},
  \bibinfo{journal}{arXiv:2402.18530}  (\bibinfo{year}{2024}).

\bibitem[{\citenamefont{Schnaus
  et~al.}(2024{\natexlab{b}})\citenamefont{Schnaus, Palackal, Poggel, Runge,
  Ehm, Lorenz, and Mendl}}]{schnaus}
\bibinfo{author}{\bibfnamefont{M.}~\bibnamefont{Schnaus}},
  \bibinfo{author}{\bibfnamefont{L.}~\bibnamefont{Palackal}},
  \bibinfo{author}{\bibfnamefont{B.}~\bibnamefont{Poggel}},
  \bibinfo{author}{\bibfnamefont{X.}~\bibnamefont{Runge}},
  \bibinfo{author}{\bibfnamefont{H.}~\bibnamefont{Ehm}},
  \bibinfo{author}{\bibfnamefont{J.~M.} \bibnamefont{Lorenz}},
  \bibnamefont{and} \bibinfo{author}{\bibfnamefont{C.~B.} \bibnamefont{Mendl}},
  \bibinfo{journal}{arXiv:2404.05448}  (\bibinfo{year}{2024}{\natexlab{b}}).

\bibitem[{\citenamefont{F.Gray}(1953)}]{Gray}
\bibinfo{author}{\bibnamefont{F.Gray}}, \bibinfo{journal}{US patent}
  \textbf{\bibinfo{volume}{2}}, \bibinfo{pages}{058} (\bibinfo{year}{1953}).

\bibitem[{\citenamefont{Mannila}(1985)}]{presortedness}
\bibinfo{author}{\bibfnamefont{H.}~\bibnamefont{Mannila}},
  \bibinfo{journal}{IEEE Transactions on Computers}
  \textbf{\bibinfo{volume}{c-34}} (\bibinfo{year}{1985}).

\bibitem[{\citenamefont{Barth et~al.}(2004)\citenamefont{Barth, Mutzel, and
  J{\"u}nger}}]{bipartite}
\bibinfo{author}{\bibfnamefont{W.}~\bibnamefont{Barth}},
  \bibinfo{author}{\bibfnamefont{P.}~\bibnamefont{Mutzel}}, \bibnamefont{and}
  \bibinfo{author}{\bibfnamefont{M.}~\bibnamefont{J{\"u}nger}},
  \bibinfo{journal}{Journal of Graph Algorithms and Applications}
  \textbf{\bibinfo{volume}{8}}, \bibinfo{pages}{179} (\bibinfo{year}{2004}).

\end{thebibliography}
\end{document}